\documentclass[letterpaper, 10 pt, conference]{ieeeconf}  

\usepackage{graphics} 
\usepackage{epsfig} 
\usepackage{mathptmx} 
\usepackage{times} 
\usepackage{amsmath} 
\usepackage{amssymb}  
\usepackage{multirow}
\usepackage{graphicx}
\usepackage{amssymb}

\usepackage{multirow}
\usepackage[table,xcdraw]{xcolor}
\usepackage{amsmath}
\usepackage[misc]{ifsym}
\usepackage{bbding}
\usepackage{pifont}
\usepackage{adjustbox}
\usepackage{booktabs}
\usepackage{subcaption}
\usepackage{graphicx}
\usepackage{booktabs}
\usepackage{adjustbox}
\usepackage{subcaption} 
\usepackage{caption}  

\IEEEoverridecommandlockouts                              

\title{\LARGE \bf
SurgFed: Language-guided Multi-Task Federated Learning for \\ Surgical Video Understanding
}

\author{Zheng Fang$^{1,2}$, Ziwei Niu$^{1,3}$, Ziyue Wang$^{1}$, Zhu Zhuo$^{1}$, Haofeng Liu$^{1}$, \\ Shuyang Qian$^{4}$, Jun Xia$^{2}$, and Yueming Jin$^{1, \dagger}$%
\thanks{$^{\dagger}$Corresponding author: Yueming Jin (ymjin@nus.edu.sg)}%
\thanks{$^{1}$Zheng Fang, Ziwei Niu, Ziyue Wang, Zhu Zhuo, Haofeng Liu, and Yueming Jin are with the National University of Singapore, Singapore (e-mail: \{niuziwei, ymjin\}@nus.edu.sg).}%
\thanks{$^{2}$Zheng Fang and Jun Xia are with the Spectral AI Team, The Hong Kong University of Science and Technology (Guangzhou), China (e-mail: \{zfang723, junxia\}@connect.hkust-gz.edu.cn).}%
\thanks{$^{3}$Ziwei Niu is also with Zhejiang University, China.}%
\thanks{$^{4}$Shuyang Qian is with Nanyang Technological University, Singapore.}%
}

\begin{document}

\maketitle
\thispagestyle{empty}
\pagestyle{empty}

\begin{abstract}
Robot-assisted minimally invasive surgery (RAS) promises safer, more precise, and increasingly autonomous interventions, which hinge on reliable surgical scene understanding. Enabling this at multi-sites scale calls for Multi-Task Federated Learning, but it remains underexplored in surgical video understanding because of two key challenges: (1) Tissue Diversity: Local models struggle to adapt to site-specific tissue features, limiting their effectiveness in heterogeneous clinical environments and leading to poor local predictions. (2) Task Diversity: Server-side aggregation, relying solely on gradient-based clustering, often produces suboptimal or incorrect parameter updates due to inter-site task heterogeneity, resulting in inaccurate localization. In light of these two issues, we propose SurgFed, a multi-task federated learning framework, enabling federated learning for surgical scene segmentation and depth estimation across diverse surgical types. SurgFed is powered by two appealing designs, i.e., Language-guided Channel Selection (LCS) and Language-guided Hyper Aggregation (LHA), to address the challenge of fully exploration on corss-site and cross-task. Technically, the LCS is first designed a lightweight personalized channel selection network that enhances site-specific adaptation using pre-defined text inputs, which optimally the local model learn the specific embeddings. We further introduce the LHA that employs a layer-wise cross-attention mechanism with pre-defined text inputs to model task interactions across sites and guide a hypernetwork for personalized parameter updates. Extensive empirical evidence shows that SurgFed yields improvements over the state-of-the-art methods in five public datasets across four surgical types. 

\end{abstract}

\section{INTRODUCTION}
Robot-assisted minimally invasive surgery (RAS) has become increasingly essential in recent decades, owing to its high mechanical stability, augmented dexterity, and growing intelligence \cite{guthart2000intuitive,long2021relational}.
Surgical scene understanding involves multiple high-level semantic downstream tasks that are crucial for enhancing the capabilities of RAS.
Multi-task learning (MTL) that simultaneously handles these tasks can benefit the context awareness of surgical scenes, including identifying the types and positions of instruments and tissues, as well as providing high-dimensional visualization of relative depth\cite{alabi2025multitask}. This not only provides cognitive assistance to surgeons \cite{endovis2018,endovis2017,loftus2020artificial} but also lays the foundation for higher-level downstream tasks, such as instrument pose estimation, tracking, task automation, and navigation \cite{pose_est,track,nagy2019dvrk}.

\begin{figure}[t]
    \centering
    \includegraphics[width=\columnwidth]{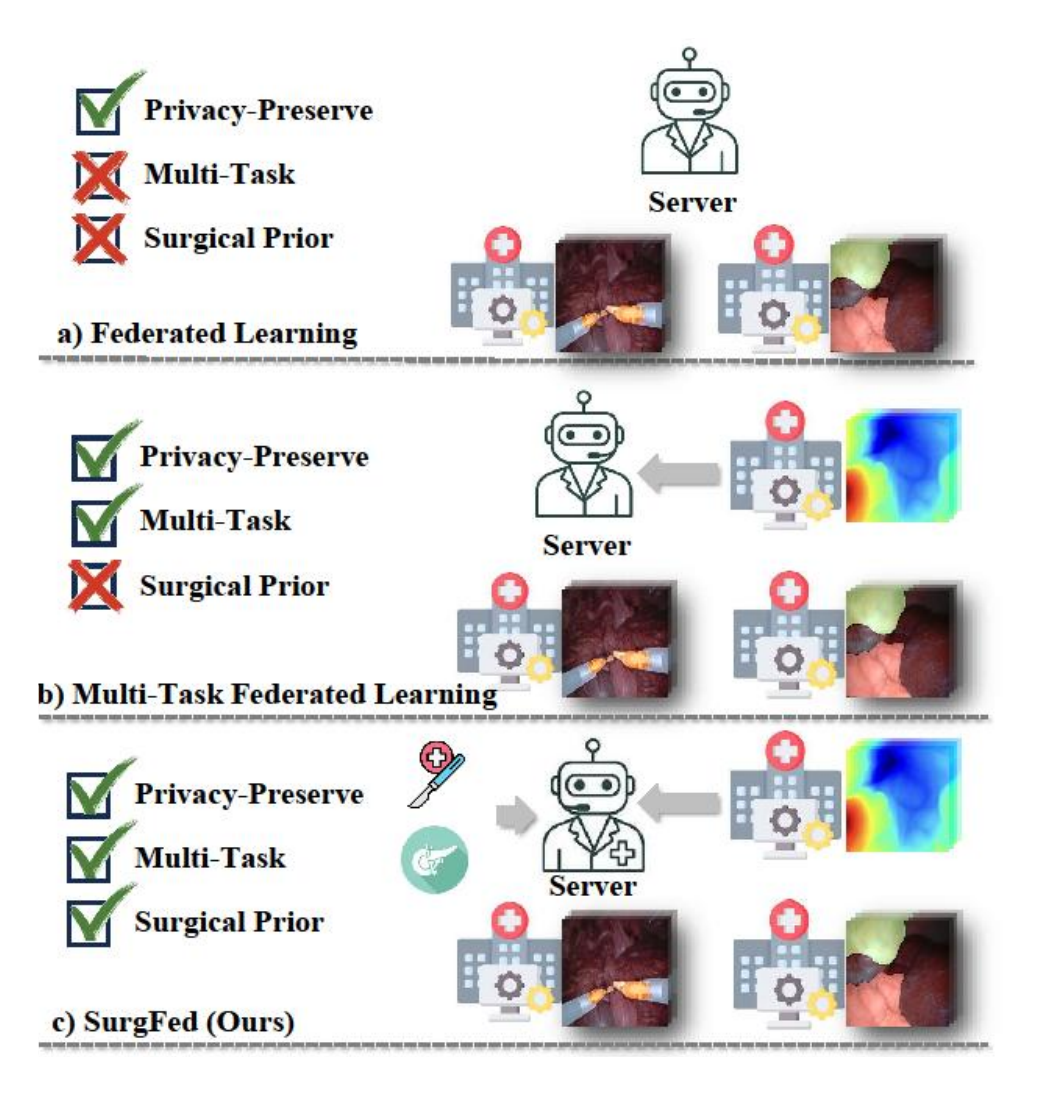}
    \caption{Comparison of different federated learning paradigms. Traditional FL fails to handle diverse surgical tasks, while Multi-Task FL supports multiple objectives but lacks domain-specific guidance. Our proposed SurgFed incorporates both multi-task capability and surgical prior, enabling personalization across different surgical scenarios.}
    \label{dif}
\end{figure}

Despite the deep learning model has shown promise in MTL, as a data-driven approach, it heavily relies on the data quantities to facilitate its efficacy \cite{he2016deep}. Collaborative training using different RAS datasets from multiple clinical sites could benefit from maximizing its potential in surgical video understanding tasks. 
Direct data communication over multiple sites is however infeasible due to privacy protection for patients and data transfer limitations. Federated Learning (FL) has been an important topic, enabling practical collaborative training without data sharing and privacy concerns \cite{kaissis2020secure}. Although FL has been widely explored, its application in surgery remains limited. To date, PFedSIS \cite{xu2025personalizing} and FedST\cite{fang2026spatio} are FL-based method for surgical instrument segmentation, introducing global-personalized disentanglement for personalized federated learning. However, it focuses solely on instrument segmentation tasks.
Beyond single-task learning, we focus on a more complex setting, Multi-Task Federated Learning (MTFL) \cite{matfl,fedlps,fedhca2,mocha}, which extends FL to handle multi-tasks through gradient-based update aggregation. 

However, \textit{MTFL in surgical videos remains under-explored}, which is due to two significant diversity challenges arising from the unique characteristics of surgical data. \textbf{\textit{{i}) Tissue Diversity}}: Due to variations in surgical types and scenes across clinical sites, surgical videos exhibit significant site-specific differences. These differences, such as diverse anatomical tissue backgrounds and surgical instruments, make it challenging for local models to effectively adapt during the MTFL process. \textbf{\textit{{ii}) Task Diversity}}: Due to varying clinical requirements across sites, task-specific label differences are significant even for the same task, further increasing inter-site heterogeneity. Traditional MTFL methods \cite{matfl,fedlps,fedhca2,mocha} that rely on gradient-based clustering struggle to handle this issue, often leading to suboptimal aggregation.

To address these challenges, we propose a multi-task federated learning scheme for surgical video understanding, namely SurgFed, which is motivated by the following two aspects: 
Firstly, to guide each site using prior surgical domain knowledge without privacy concerns, we introduce a Language-Guided Channel Selection (LCS) mechanism, which utilizes pre-defined text inputs describing surgical types, task types, and tissue characteristics to guide feature selection at each site.
In this way, we are able to realize personalized channel selection on the encoder representations for effective site-specific adaptation. Secondly, soley rely on gradient aggregation, which may overlook the semantic relationships between tasks and struggle to effectively model inter-site task dependencies. Inspired by this, we design a Language-Guided Hyper Aggregation (LHA), leveraging a layer-wise cross-attention mechanism and pre-defined text inputs to model the interaction between different tasks across different sites. This interaction serves as an indicator for a hyper-network to update the parameters of each site. In this way, we capture cross-task semantic dependencies and site-specific variations, enabling a more structured and interpretable aggregation while efficiently learning mutual information in a lightweight manner.

Our main contributions are summarized as follows:
\begin{itemize}
  \item We introduce language-guided surgical priors to federated surgical video understanding for the first time. By leveraging pre-defined textual prompts describing surgical instruments and anatomical structures, we inject semantic knowledge into model adaptation across heterogeneous surgical types and institutions.

  \item We propose a novel Language-Guided Federated Architecture, which consists of two modules: (1) Language-Guided Channel Selection (LCS), a personalized trainable adapter that selects encoder channels conditioned on text prompts for site-specific representation personalization; and (2) Language-Guided Hyper Aggregation (LHA), which introduces a task-aware, cross-site hypernetwork guided by language inputs to dynamically update model based on inter-site task similarity.

  \item We validate our method on five public surgical datasets from different clinical centers, covering four distinct surgical types and encompassing both surgical scene segmentation and depth estimation tasks. Our proposed method consistently outperforms existing state-of-arts methods, showcasing the adaptability and robustness.
\end{itemize}

\begin{figure*}[ht]
    \centering
    \includegraphics[width=0.9\linewidth]{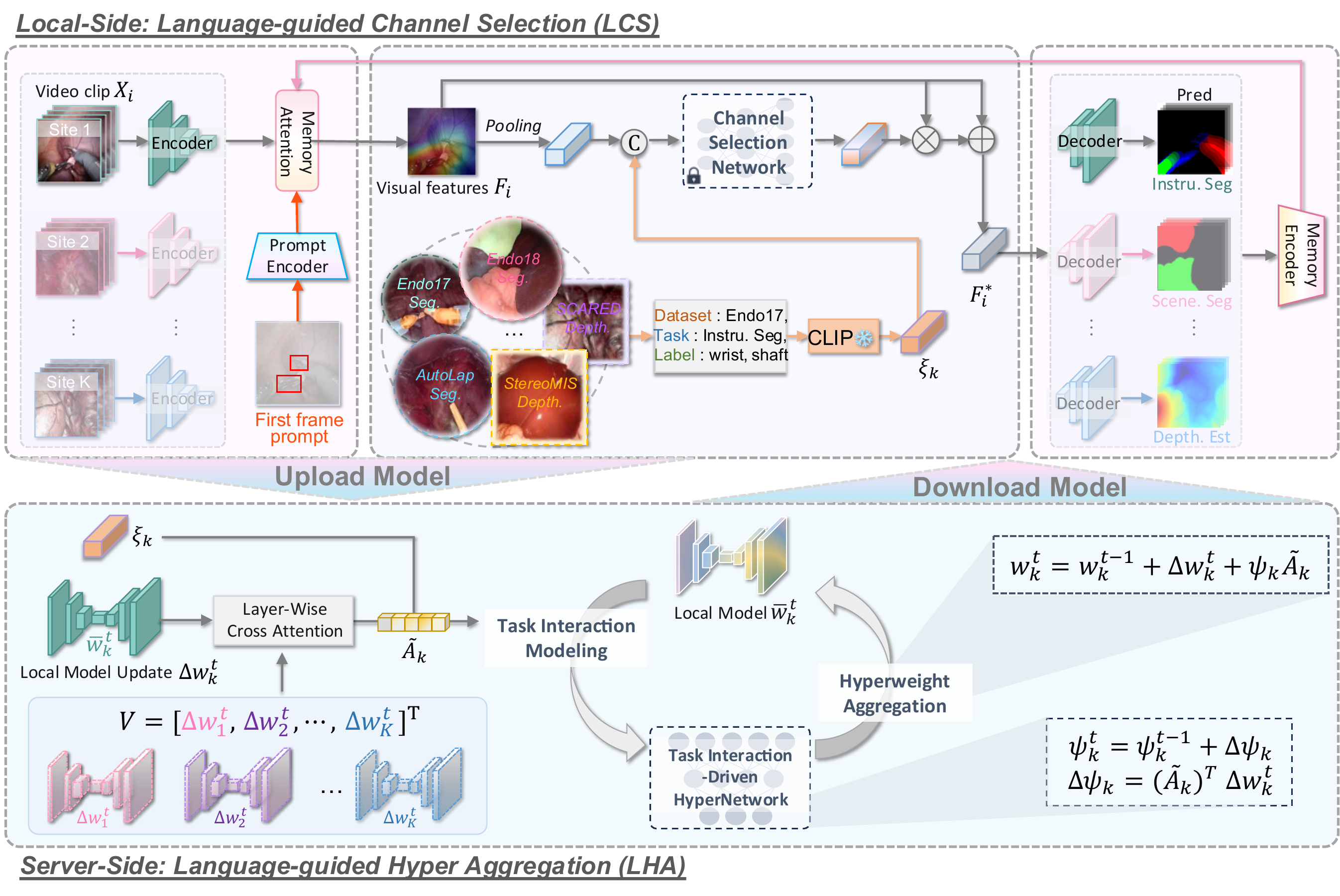}
    \caption{In the local-side stage, LCS allows each local model to adapt to local data through personalized selection and enhancement of feature-specific channels. In the server-side stage, LHA models the task interactions across different sites, enabling a personalized update to each local site model. }
    \label{method}
\end{figure*}

\section{Related Work}

\noindent \textbf{Surgical Video Understanding.}
Extensive studies have been conducted on both surgical instrument segmentation and depth estimation. For the former, techniques such as holistically-nested networks \cite{garcia2017toolnet}, graph-based networks \cite{liu2021graph}, synergistic networks \cite{wang2024video}, contrastive learning \cite{lou2023min}, active learning \cite{peng2024reducing}, and the incorporation of auxiliary cues like depth maps \cite{mohammed2019streoscennet}, optical flow \cite{jin2019incorporating}, motion flow \cite{zhao2020learning}, tracking signals \cite{zhao2022trasetr}, and synthetic images \cite{colleoni2021robotic} have been explored. Recently, the emergence of the Segment Anything Model (SAM) \cite{kirillov2023segment} has sparked new interest in universal segmentation. Efforts such as SurgSAM \cite{surgsam, liu2025resurgsam2, liu2025sam2s, liu2024surgical} have adapted SAM to the surgical domain.

For the latter, depth estimation approaches include stereo-based methods \cite{reiter2016endoscopic, allan2021stereoscopic} and monocular prediction \cite{hu2020monocular, luo2022geosurgical}, with the latter gaining traction due to its practicality. Depth provides valuable geometric priors that enhance segmentation \cite{luo2022geosurgical} and support multi-task learning \cite{wang2023twin}, self-supervised estimation \cite{ye2022self, chen2023unsupervised}, and domain adaptation. Nonetheless, existing works focus on isolated surgical types or datasets, lacking collaborative training across domains.

\noindent \textbf{Federated Learning and Multi-Task Federated Learning.} 
Owing to data privacy concerns, FL has attracted increasing attention in fields, especially in the medical domain, as it enables decentralized model training across sites without sharing raw data and has achieved notable success \cite{fedavg,silva2019federated,guan2024federated,subedi2023client}.
FL has been applied to brain tumor segmentation by using FedAvg to combine insights from different centers \cite{li2019privacy}. It has also been utilized for detecting COVID-19 anomalies in lung CT scans, incorporating semi-supervised learning to boost performance \cite{dou2021federated}. FL faces limitations due to data heterogeneity, such as differing medical imaging devices and protocols \cite{yu2022salvaging}.
To mitigate these issues, PFL has been proposed to allow partial model parameter personalization for each local site, so that the federated model can be adapted to distinct data distribution across different sites. PFL divides model parameters into global and personalized layers, such as prediction heads \cite{fedrep}, batch normalization \cite{li2021fedbn}, convolution channels \cite{shen2022cd2}, or query embeddings in self-attention \cite{feddp}. Additionally, server-side optimization approaches, such as weighting sites using the Fisher Information Matrix \cite{yang2024fedas} and decomposing uploaded gradients \cite{zeng2024tackling}, have been explored to further enhance PFL's adaptability to heterogeneous data.
Despite advancements in medical applications, investigations into applying FL and PFL for surgical instrument segmentation remain limited. Existing FL methods often overlook key characteristics of the surgical domain, such as the intricate temporal nature of video data, resulting in suboptimal performance and underscoring the need for further research.

\section{Methodology}

\subsection{Overall Pipeline} 
\textbf{Problem Formulation.} Assume that there are $K$ local sites, each with a unique non-iid distribution $\mathcal{D}_k$ over a set of $N_k$ samples $\left\{\left(x_{i,k}, y_{i,k}\right)\right\}_{i=1}^{N_k}$, where $\left(x_{i,k}, y_{i,k}\right) \in \mathcal{D}_k$. The joint image and label space across these sites is denoted as $\left\{\left(\mathcal{X}_k, \mathcal{Y}_k\right)\right\}_{k=1}^K$.
MTFL methods propose to learn the respective local models $\{w_k^t\}_{k=1}^K$ for all local sites in the communication round $t$ with task $\Gamma_i$, where each local model $w_k$ aims to suit the distribution of the $k$-th site:
\begin{equation}
\label{pfl}
w_k^{t+1} \;\boldsymbol{=}\; w_k^t \;\boldsymbol{-}\; \eta \, \mathbb{E}_{(x_{i,k},\, y_{i,k}) \in \mathcal{D}_k} \nabla \mathcal{L}_k\!\big(f_{w_k}(x_{i,k}),\, y_{i,k}\big),
\end{equation}
where $\eta$ is the learning rate; and $\mathcal{L}_k$ is the $k$-th site's loss function.
At each site $k$, the local objective is to perform pixel-wise prediction over video clips $\{\boldsymbol{X}_{i,k}, \boldsymbol{Y}_{i,k}\}_{i=1}^{N_k}$, where $\boldsymbol{X}_{i,k}$ denotes the input frames and $\boldsymbol{Y}_{i,k}$ is the corresponding ground truth, either in the form of semantic segmentation masks or depth maps. The loss function $\mathcal{L}_k$ is task-specific.
For segmentation tasks, we adopt the standard pixel-wise cross-entropy loss:
\[
\mathcal{L}_k^{\text{seg}} \;\boldsymbol{=}\; \boldsymbol{-} \; \boldsymbol{\sum}_{c=1}^{C_k} 
y_{i,k}^{(c)} \log f_{w_k}^{(c)}(x_{i,k}),
\]
where $C_k$ is the number of semantic classes at site $k$, $y_{i,k}^{(c)}$ is the one-hot label for class $c$, and $f_{w_k}^{(c)}$ denotes the predicted probability for class $c$.
For depth estimation tasks, we use the L1 norm loss (mean absolute error), which is commonly adopted in depth estimation due to its robustness to outliers and ability to preserve structural details:
\[
\mathcal{L}_k^{\text{depth}} \;\boldsymbol{=}\; 
\left\| f_{w_k}(x_{i,k}) \;\boldsymbol{-}\; y_{i,k} \right\|_1 .
\]

\noindent\textbf{Overview of SurgFed.} Our SurgFed provides a MTFL framework to achieve surgical video scene segmentation and depth estimation in a decentralized manner. It performs in two stages as Fig.~\ref{method} shows: 
In the local-side stage, LCS (refer to Sec. \ref{LCS}) allows each local model to adapt to local data through personalized selection and enhancement of feature-specific channels.
In the server-side stage, LHA (refer to Sec. \ref{lha}) models the task interactions across different sites, enabling a personalized update to each local site model.
We perform $T$ rounds of communication, with $E$ local updates per round. Local models are optimized using LCS, while the server applies LHA each round by aggregating updated parameters from all sites. After $T$ rounds, personalized models are obtained without disclosing any local data.

\subsection{Language-guided Channel Selection (LCS)}\label{LCS}
Due to variations in surgical protocols and surgery types, each site exhibits unique tissue backgrounds and significant differences in ground truth labels, requiring adaptation to intra-site knowledge such as site-specific labels and surgical contexts. We incorporate a personalized lightweight channel selection network, which is not shared in the FL process, and a pre-trained CLIP \cite{clip} model with predefined text to represent intra-site surgical knowledge, enhancing visual representation by highlighting task-relevant channels. However, as CLIP is pre-trained on general-domain data, it lacks surgical context awareness. Our predefined text and channel selection network mitigate this limitation by guiding CLIP to capture site-specific differences and task-relevant features.

We employ a SAM2-based \cite{medsam2} architecture to process surgical videos. The visual features extracted from each video clip are denoted as \(\boldsymbol{F}_i\), where \(\boldsymbol{F} \in \mathbb{R}^{l \times h \times w \times c}\), with \(l\), \(h\), \(w\), and \(c\) representing the length, height, width, and number of channels, respectively.
We utilize predefined text prompts, such as \textit{``Dataset: EndoVis2017, Task: Instrument Segmentation, Label: Shaft, Wrist, Clasper”}, as input for the pre-trained CLIP model, generating a one-dimensional text embedding \(\xi_k \in \mathbb{R}^d\). This embedding is then extended along the spatial dimensions using the \texttt{expand} operation, resulting in \(\xi_k \in \mathbb{R}^{l \times h \times w}\), which contains intra-site knowledge of the data from the $k$-th site.

To generate a composite indicator that enables dynamic channel selection for each feature, we first perform the global average pooling on \(\boldsymbol{F}_{i} \in \mathbb{R}^{l \times h \times w \times c}\). These pooled values are then integrated with the text embedding \(\xi_k \in \mathbb{R}^{l \times h \times w}\) via concatenation operation $[\cdot,\cdot]$. The augmented feature is then fed into a fully connected layer followed by a sigmoid activation \(\sigma\), resulting in the composite indicator \(\hat{\xi}_k \in \mathbb{R}^{l \times h \times w}\). This process is expressed as:  
\begin{equation}
    \label{hat}
    \hat{\xi}_k \;\boldsymbol{=}\; 
    \sigma\!\left(\mathrm{FC}\big([\text{AvgPool}(\boldsymbol{F}_{i}),\, \text{Expand}(\xi_k)]\big)\right).
\end{equation} 
The composite indicator \(\hat{\xi}_k\) is designed to selectively enhance relevant channels by integrating both the text embedding and the visual features. This enhancement is achieved by applying the composite indicator \(\hat{\xi}_k\) to the feature map \(\boldsymbol{F}_{i}\) through pixel-wise multiplication, while a residual connection ensures robustness against potential errors in channel selection. This process is formulated as \[
\boldsymbol{F}_{i}^{*} \;\boldsymbol{=}\; 
\boldsymbol{F}_{i} \;\boldsymbol{+}\; 
\boldsymbol{F}_{i} \;\boldsymbol{\otimes}\; \hat{\xi}_k ,
\]
where \(\otimes\) denotes pixel-wise multiplication. Finally, the enhanced features \(\boldsymbol{F}_{i}^*\) are fed into the decoder to generate the segmentation mask or depth map.

\subsection{Language-guided Hyper Aggregation (LHA)}
\label{lha}
With each local model adapted to its intra-site knowledge through LCS, exploring inter-site knowledge, such as similar instrument structures and motions, is crucial for collaborative training. Gradient-based clustering, commonly used for aggregation, struggles in the surgical domain due to diverse task labels, often leading to suboptimal aggregation.
Thus, we propose to explore the inter-site knowledge by leveraging both the model updates (i.e., gradients) and pre-defined text inputs that describe the specific tasks at each site. A task interaction-driven hypernetwork on the server side is trained with text and gradient as indicators, enabling the exploration of task interactions across sites. 

Formally, we denote the model parameters for the \(k\)-th site at round \(t\) after applying the local LCS update as \(\overline{w}_k^t\). Thus, the gradient update $\Delta w_k^t$ is given by: $\Delta w_k^t = \overline{w}_k^t - w_k^{t-1}$. Then, we model the cross-task interaction via layer-wise cross-attention mechanism as:
\begin{equation}
    V = [\Delta w_1^t,\Delta w_2^t,...,\Delta w_k^t]^\mathsf{T},
\end{equation}
\begin{equation}
     A_{k} = \text{Softmax}(\Delta w_k^tV^{\mathsf{T}}/ \sqrt{d})/V,
\end{equation}
where \(V\) represents the stacked gradient updates from all \(K\) local sites with a dimension of \(d\); \(A_{k}\) is the attention matrix for the \(k\)-th site. Relying solely on gradient information for updates still lacks sufficient guidance such as higher-level task similarities and differences. Thus, we incorporate the same pre-defined text prompts from pre-trained CLIP to emphasize the inter-site knowledge as:
\begin{equation}
    \xi^{*}_k \;\boldsymbol{=}\; 
    \sigma\!\left(\mathrm{FC}\big([\text{AvgPool}(A_{k}),\, \text{Expand}(\xi_k)]\big)\right),
\end{equation}
\begin{equation}
     \tilde{A_{k}} \;\boldsymbol{=}\; 
     A_{k} \;\boldsymbol{+}\; A_{k} \;\boldsymbol{\otimes}\; \xi^{*}_k ,
\end{equation}
Unlike the previous channel-focused feature enhancement in LCS, this language-guided indicator is designed to highlight inter-site relationships by emphasizing which sites and layers contribute more significantly to the aggregation process.
To dynamically adjust aggregation based on inter-site relationships, we propose to exploit a hypernetwork on the server side to update each local site, which assigns adaptive importance to each site's updates. The personalized aggregation for the \(k\)-th site model which will be uploaded to server next round is formulated as:
\begin{equation}
    w_k^t \;\boldsymbol{=}\; 
    w_k^{t-1} \;\boldsymbol{+}\; \Delta w_k^t \;\boldsymbol{+}\; \psi_k \tilde{A}_k ,
\end{equation}
where $\psi_k$ is a learnable weight and can be derived by the chain rule: $\nabla_{\psi_k} \mathcal{L}_k = (\tilde{A}_k)^\mathsf{T} \nabla_{w_k^t} \mathcal{L}_k.$ Thus, $\psi_k$ can be updated during FL process as: 
\begin{equation}
    \psi_k^{t} \;\boldsymbol{=}\; \psi_k^{t-1} \;\boldsymbol{+}\; \Delta \psi_k, 
    \quad \Delta \psi_k \;\boldsymbol{=}\; (\tilde{A}_k)^\mathsf{T} \Delta w_k^t .
\end{equation}

\begin{table*}[t]
\centering
\caption{Comparison of state-of-the-art methods across multiple metrics. Dice(\%) and IoU(\%) for segmentation, RMSE (pix.) for monocular depth estimation, and $\Delta m$ (\%) as the overall metric.}
\label{maintab}
\footnotesize 
\renewcommand{\arraystretch}{1.3} 
\setlength{\tabcolsep}{3pt} 

\resizebox{\textwidth}{!}{%
\begin{tabular}{l|ccc|ccc|ccc|cc|cc|c}
\toprule
\textbf{Method} & \multicolumn{3}{c|}{\textbf{Endovis2017}} & \multicolumn{3}{c|}{\textbf{Endovis2018}} & \multicolumn{3}{c|}{\textbf{AutoLaparo}} & \multicolumn{2}{c|}{\textbf{SCARED}} & \multicolumn{2}{c|}{\textbf{StereoMIS}} & \textbf{Avg} \\
 & IoU$\uparrow$ & Dice$\uparrow$ & $\Delta m$$\uparrow$ & IoU$\uparrow$ & Dice$\uparrow$ & $\Delta m$$\uparrow$ & IoU$\uparrow$ & Dice$\uparrow$ & $\Delta m$$\uparrow$ & RMSE$\downarrow$ & $\Delta m$$\uparrow$ & RMSE$\downarrow$ & $\Delta m$$\uparrow$ & $\Delta m$$\uparrow$ \\ \midrule

Inference Only & 44.60$_{\pm0.00}$ & 54.17$_{\pm0.00}$ & $-$23.62$_{\pm0.00}$ & 63.31$_{\pm0.00}$ & 71.77$_{\pm0.00}$ & $-$10.92$_{\pm0.00}$ & 80.63$_{\pm0.00}$ & 87.02$_{\pm0.00}$ & $-$1.65$_{\pm0.00}$ & --- & --- & --- & --- & --- \\
Local Train & 58.77$_{\pm0.74}$ & 70.47$_{\pm1.28}$ & 0.00$_{\pm0.42}$ & 71.53$_{\pm0.79}$ & 80.06$_{\pm0.91}$ & 0.00$_{\pm0.20}$ & 82.39$_{\pm0.82}$ & 88.04$_{\pm1.66}$ & 0.00$_{\pm0.37}$ & 10.76$_{\pm0.23}$ & 0.00$_{\pm0.69}$ & 14.75$_{\pm0.54}$ & 0.00$_{\pm0.13}$ & 0.00$_{\pm0.71}$ \\
FedAvg & 54.59$_{\pm0.53}$ & 66.38$_{\pm0.79}$ & $-$6.46$_{\pm0.71}$ & 65.06$_{\pm0.29}$ & 74.21$_{\pm0.03}$ & $-$8.18$_{\pm0.43}$ & 83.19$_{\pm0.76}$ & 88.59$_{\pm0.25}$ & 0.80$_{\pm0.35}$ & 28.61$_{\pm0.71}$ & $-$165.75$_{\pm0.64}$ & 14.77$_{\pm0.38}$ & $-$0.17$_{\pm0.74}$ & $-$35.95$_{\pm1.56}$ \\
FedAvg+Cluster & 57.21$_{\pm0.57}$ & 69.44$_{\pm0.15}$ & $-$2.06$_{\pm0.32}$ & 65.76$_{\pm1.08}$ & 75.51$_{\pm1.46}$ & $-$6.87$_{\pm0.30}$ & 83.26$_{\pm0.45}$ & 88.71$_{\pm0.50}$ & 0.91$_{\pm2.00}$ & 34.09$_{\pm0.61}$ & $-$216.67$_{\pm0.60}$ & 15.71$_{\pm0.40}$ & $-$6.57$_{\pm0.46}$ & $-$46.25$_{\pm0.37}$ \\
FedRep & 58.59$_{\pm0.45}$ & 70.58$_{\pm1.04}$ & $-$0.08$_{\pm1.34}$ & 70.94$_{\pm0.38}$ & 79.32$_{\pm0.59}$ & $-$0.87$_{\pm0.43}$ & 83.37$_{\pm0.09}$ & 88.86$_{\pm0.67}$ & 1.06$_{\pm1.45}$ & 16.78$_{\pm0.31}$ & $-$55.86$_{\pm0.78}$ & 14.63$_{\pm0.98}$ & 0.77$_{\pm0.50}$ & $-$10.99$_{\pm0.92}$ \\
FedProx & 53.22$_{\pm0.78}$ & 64.85$_{\pm0.77}$ & $-$8.71$_{\pm0.11}$ & 67.35$_{\pm1.86}$ & 76.28$_{\pm1.09}$ & $-$5.28$_{\pm0.77}$ & \textbf{83.66$_{\pm0.33}$} & \textbf{89.06$_{\pm0.05}$} & \textbf{1.35$_{\pm2.00}$} & 128.59$_{\pm0.86}$ & $-$1094.62$_{\pm0.60}$ & 16.20$_{\pm0.50}$ & $-$9.88$_{\pm0.86}$ & $-$223.43$_{\pm0.60}$ \\ \hline
\rowcolor{gray!10} MaT-FL & 56.19$_{\pm1.06}$ & 67.80$_{\pm0.23}$ & $-$4.09$_{\pm1.64}$ & 71.01$_{\pm1.30}$ & 79.48$_{\pm0.22}$ & $-$0.73$_{\pm0.77}$ & 83.27$_{\pm0.19}$ & 88.75$_{\pm0.50}$ & 0.94$_{\pm2.00}$ & 10.58$_{\pm0.56}$ & 1.67$_{\pm0.54}$ & 15.38$_{\pm1.21}$ & $-$4.29$_{\pm1.73}$ & $-$1.30$_{\pm0.65}$ \\
\rowcolor{gray!10} FedHCA$^2$ & 59.73$_{\pm0.71}$ & 71.27$_{\pm0.42}$ & 1.38$_{\pm0.89}$ & 70.25$_{\pm0.74}$ & 78.67$_{\pm0.50}$ & $-$1.76$_{\pm0.85}$ & 83.61$_{\pm0.40}$ & 89.03$_{\pm1.74}$ & 1.30$_{\pm1.91}$ & 10.25$_{\pm0.71}$ & 4.79$_{\pm1.04}$ & 18.81$_{\pm0.38}$ & $-$27.57$_{\pm2.00}$ & $-$4.37$_{\pm0.37}$ \\
\rowcolor{gray!25} \textbf{SurgFed (Ours)} & \textbf{62.17$_{\pm0.52}$} & \textbf{73.76$_{\pm0.15}$} & \textbf{5.23$_{\pm0.30}$} & \textbf{73.33$_{\pm0.13}$} & \textbf{81.44$_{\pm0.13}$} & \textbf{2.12$_{\pm0.14}$} & 83.45$_{\pm0.61}$ & 88.89$_{\pm0.17}$ & 1.13$_{\pm0.35}$ & \textbf{8.78$_{\pm0.22}$} & \textbf{18.42$_{\pm1.32}$} & \textbf{14.34$_{\pm0.38}$} & \textbf{2.73$_{\pm1.09}$} & \textbf{5.92$_{\pm0.48}$} \\ 
\bottomrule
\end{tabular}
}
\end{table*}

\section{Experiments}

\begin{table*}[h]
\centering
\caption{Quantitative analysis of the key components across five sites, with values in parentheses indicating differences from the baseline (FedAvg).}
\label{ablation}
\resizebox{1\linewidth}{!}{%
\begin{tabular}{c|ccc|ccc|cc|c}
\toprule
\# & LCS & LHA & \begin{tabular}[c]{@{}c@{}} Text\\ Pro. \end{tabular}  
& \begin{tabular}[c]{@{}c@{}}Endovis2017 \\ Dice $\uparrow$ \end{tabular}  
& \begin{tabular}[c]{@{}c@{}}Endovis2018 \\ Dice $\uparrow$ \end{tabular}  
& \begin{tabular}[c]{@{}c@{}}AutoLaparo \\ Dice $\uparrow$ \end{tabular}  
& \begin{tabular}[c]{@{}c@{}}SCARED \\ RMSE $\downarrow$\end{tabular}  
& \begin{tabular}[c]{@{}c@{}}StereoMIS \\ RMSE $\downarrow$ \end{tabular}  
& \begin{tabular}[c]{@{}c@{}} Avg. \\ $\Delta m \uparrow$ \end{tabular} \\
\hline
  1 & --- &  ---  &   ---   & 66.38(0.00)        & 74.21(0.00)      & 88.59(0.00) & 28.61(0.00) & 14.77(0.00)     & $-35.95(0.00)$  \\
2 & \ding{51}   &  ---   &     ---  & $73.02(+6.64)$ & $81.08(+6.87)$    & $88.73(+0.14)$  & $10.91(-17.70)$   & $14.39(-0.38)$      & $1.50(+37.45)$   \\
   3 & --- & \ding{51}   &         ---            & $72.12(+5.74)$ & $79.84(+5.63)$   & $\textbf{89.01(+0.42)}$       & $\textbf{8.77(-19.84)}$  & $14.64(-0.13)$      & $4.54(+40.49)$  \\
   4 & ---  & \ding{51}   & \ding{51}           & $\textbf{74.59(+8.21)}$        & $81.27(+7.06)$       & $88.93(+0.34)$       & $8.84(-19.77)$  & $14.49(-0.28)$      & $5.84(+41.79)$ \\
5 & \ding{51}   & \ding{51}   &  ---  & $73.76(+7.38)$        & $\textbf{81.44(+7.23)}$      & $88.89(+0.30)$       & $8.78(-19.83)$  & $\textbf{14.34(-0.43)}$      & $\textbf{5.92(+41.87)}$ \\
\bottomrule
\end{tabular}
}
\end{table*}

\begin{figure*}[h]
    \centering
    \includegraphics[width=\linewidth]{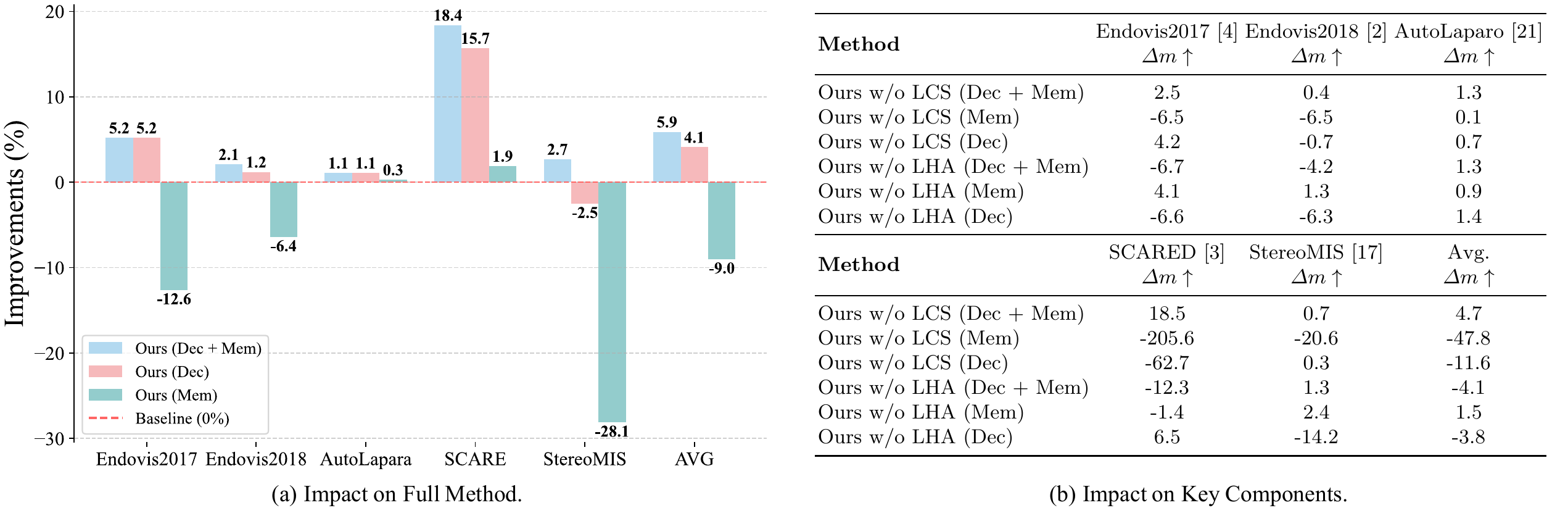}
    \caption{Impact on Fine-Tuning Dec and Mem Layers of SAM2.}
    \label{finetune}
\end{figure*}

\begin{figure*}[t]
    \centering
    \includegraphics[width=0.8\linewidth]{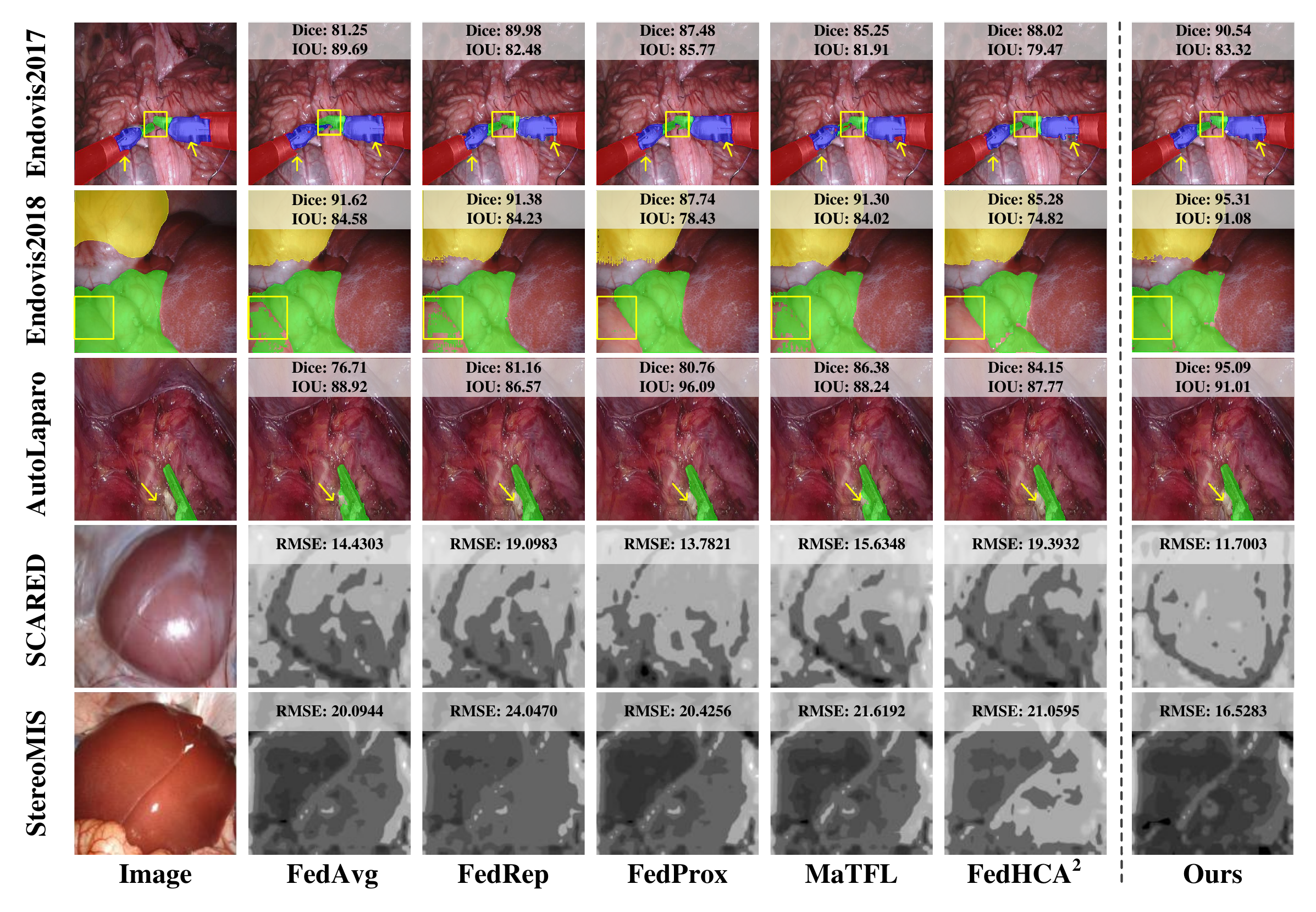}
    \caption{Segmentation and depth estimation visualization results across five sites.}
    \label{vis}
    \vspace{5mm}
\end{figure*}

\subsection{Experiment Setting}
\textbf{Datasets and Evaluation.}
We construct a new benchmark for MTFL in the surgical domain using five public datasets for method evaluation and comparison.
1) EndoVis2017 dataset \cite{endovis2017} focuses on surgical instrument part segmentation with 3 classes from Nephrectomy procedures.  
2) EndoVis2018 dataset \cite{endovis2018} extends segmentation to the whole surgical scene, covering 11 classes within the same surgical type.  
3) AutoLaparo dataset \cite{autolaparo} provides surgical scene segmentation with 9 classes for Hysterectomy procedures.  
4) SCARED dataset \cite{scared} is designed for depth estimation in porcine cadaver abdominal anatomy.  
5) StereoMIS dataset \cite{stereomis} consists of endoscopic videos from in-vivo porcine subjects for depth estimation. Following \cite{cui2024endodac,surgsam2,fedhca2}, we use Dice and IoU for the evaluation of segmentation, RMSE for monocular depth estimation, and the percentage of improvement $\Delta m $ over local training setting as the overall performance metric:
\begin{equation}
\Delta m = \frac{1}{K} \sum_{k=1}^{K} 
s_k \cdot \frac{M_k - M_k^{\text{local}}}{M_k^{\text{local}}} \times 100\% ,
\end{equation}
where $K$ is the number of evaluation metrics, $M_k$ denotes the result on the $k$-th metric, $M_k^{\text{local}}$ is the corresponding local training baseline, and $s_k \in \{+1,-1\}$ indicates whether the metric is ``higher-is-better'' ($+1$) or ``lower-is-better'' ($-1$).

\noindent \textbf{Implementation Details.} 
We employ the SAM2 architecture as the backbone and exploit the data augmentation strategy, bounding-box prompts, and training strategies as described in \cite{medsam2}. During inference, we provide only the first frame's bounding box. For the segmentation task, this simulates the surgeon's instruction regarding the region of interest at the start of the surgery; for depth estimation, we utilize a whole-image-size bounding box as the prompt. 
To adapt the original SAM2 for multi-task surgical intelligence, we implement a decoupled architectural strategy. While the image encoder and memory attention modules serve as a shared global backbone, we extend the decoder section into task-specific branches. Specifically, we maintain the transformer-based mask decoder for segmentation and append a task-specific regression head for depth estimation. To facilitate seamless migration across these tasks, we utilize the standard SAM2 as a unified foundation. Our primary contribution is a general MTFL framework rather than a task-specific model specialized for a single modality, which justifies the use of a versatile backbone over task-specific state-of-the-art models.
We freeze the encoder while fine-tuning the memory and decoder layers. For local training, we use the Adam optimizer with a learning rate of $1\mathrm{e}^{-4}$ and $E=3$ local epochs per round. The server-side hypernetwork is optimized using the Adam optimizer with a learning rate of $1\mathrm{e}^{-3}$, with a total of $T=100$ communication rounds. To demonstrate the effectiveness of SurgFed, we compare it with five state-of-the-art methods (Table~\ref{maintab}), including FL methods (FedAvg~\cite{fedavg}, FedRep~\cite{fedrep}, FedProx~\cite{fedprox}), FedAvg+Cluster (task-specific aggregation), and MTFL methods (MaT-FL, FedHCA$^2$).

\subsection{Experimental Results and Empirical Analysis}
To demonstrate the effectiveness of SurgFed, we compare it with five state-of-the-art methods (Table~\ref{maintab}), including the FL methods FedAvg~\cite{fedavg}, FedRep~\cite{fedrep}, and FedProx~\cite{fedprox}, as well as FedAvg+Cluster, which adopts task-specific aggregation by separately aggregating segmentation and depth estimation. We also compare with the MTFL methods MaT-FL and FedHCA$^2$.
We also perform local training (Local Train) as our baseline, where each model is trained using only its own dataset without FL. Furthermore, we include a pre-trained SAM2 model evaluated in an inference-only setting (Inference Only) without any fine-tuning. Note that we primarily focus on FL methods capable of handling heterogeneous task objectives. 
While recent FL variants such as~\cite{li2021fedbn,shen2022cd2,feddp,yang2024fedas} do not inherently support the multi-task setting (e.g., simultaneous segmentation and depth estimation) where architectural branch-outs and non-overlapping label spaces are present, making them unsuitable for direct comparison in this study.

The SAM2-based model demonstrates strong zero-shot capabilities but performs poorly when directly applied to surgical tasks without fine-tuning, especially for non-segmentation tasks. This highlights its limited generalizability in the surgical domain.
FL methods show improvements in certain sites, confirming the benefits of collaborative training. However, these methods fail to consider the inherent task diversity across sites, leading to suboptimal global aggregation and performance drop (FedRep -10.99\%). 
While multi-task FL methods better capture both task similarities and differences across sites, they still do not fully account for the characteristics of surgical data, such as highly diverse tissue backgrounds (FedRep -10.99\% vs. MaT-FL -1.30\%). As a result, local site models struggle to effectively focus on their own data and tasks. Moreover, the lack of high-level guidance in the aggregation process leads to incorrect parameter updates, further limiting performance improvements.
Our method leverages LHA to model task interactions across sites through language-guided representations, enabling accurate aggregation. Additionally, LCS personalizes each site by focusing on site-specific features, 
leading to consistent performance improvements across all five sites. 
Notably, FedProx excels on AutoLaparo, but under stronger domain shifts its proximal constraint impedes adaptation and—being single-task—cannot handle MTL, yielding erroneous depth-estimation predictions; accordingly, our method, tailored for high-heterogeneity MTL, shows limited gains on AutoLaparo and is not the best there.
The visualization results are shown in Fig~\ref{vis}.

\subsection{Ablation Study}
\textbf{Effectiveness of Key Components.} 
We conduct ablation experiments to validate the effectiveness of different key components in the proposed method and obtain five configurations. 
The results are presented in Table~\ref{ablation}: \textbf{1st row}: we train the pure FedAvg method as the baseline, where no LCS and LHA are involved. 
\textbf{2nd row}: LCS is utilized to personalize the representation of each site in FL. 
\textbf{3rd row}: LHA is included to model the inter-site task interaction. 
\textbf{4th row}: we include the textual prompt embedding without the channel selection network, aiming to provide site-specific guidance. 
\textbf{5th row}: our proposed full SurgFed.

As shown in Table~\ref{ablation}, an improvement of +1.50\% is achieved when applying LCS (\#2) to the baseline FedAvg (\#1). LCS enhances performance by personalizing site-specific features but still struggles with high task diversity datasets such as SCARED due to the lack of inter-site task interaction modeling. Conversely, LHA only (\#3) yields significant improvements in depth estimation by modeling inter-site task relationships but offers limited gains in segmentation, as it lacks site-specific semantic adaptation. Building on LHA, combining it with Local Text Prompts (\#4) highlights the benefits of task-specific text guidance. However, our full method (\#5) further refines feature representations, achieving a significant improvement of +5.92\%, demonstrating the importance of adaptive channel selection.



\noindent \textbf{Impact on Different Indicators in LCS and LHA.}
We evaluate the impact of different indicator designs in both LCS and LHA modules. The results are shown in Tables~\ref{lcs_indicator} and~\ref{lha_indicator}.
Notably, the performance gap is more significant in the LHA module. For example, the random initialization baseline leads to strong negative transfer on SCARED (--8.08\%) and Endo18 (--3.99\%), with an overall performance drop of $\Delta m = -1.26$. One-hot embeddings yield moderate improvements on certain datasets (e.g., +6.36\% on SCARED), but their lack of semantic continuity across sites limits generalization.
In contrast, our method achieves consistent improvements across all datasets, including challenging out-of-distribution domains like SCARED (+18.42\%) and yields the highest average gain ($\Delta m = +5.92$). This suggests that LHA, which performs cross-task aggregation (e.g., combining segmentation and depth cues), requires stronger semantic guidance to align heterogeneous task features. Simple or unstructured indicators fail to capture transferable knowledge across tasks and domains, while our semantic-aware indicator enables effective generalization by leveraging textual descriptions of surgical contexts.

\noindent \textbf{Impact on Fine-Tuning Different Layers of SAM2.}
We investigate how different components of SAM2 contribute to overall performance by fine-tuning the Decoder layers (Dec) and Memory layers (Mem), as shown in Fig.~\ref{finetune}. Specifically, Fig.~\ref{finetune}.a compares full models with different combinations of Dec and Mem tuning, while Fig.~\ref{finetune}.b reports ablation results under the absence of LCS or LHA.

Under the w/o LHA setting, fine-tuning Mem layers leads to better performance ($\Delta m$ = +1.5\%) than fine-tuning Dec layers ($\Delta m$ = --3.8\%). This suggests that memory components play a more critical role when cross-task supervision is missing. Since LHA enables task-level alignment between segmentation and depth, removing it causes a loss of shared guidance. However, Mem layers can partially recover the missing alignment by leveraging temporal or motion-related cues across frames, enabling more robust knowledge transfer.
In contrast, under the w/o LCS setting, both Mem and Dec layers are essential. Removing LCS eliminates cross-site personalization, causing the model to generalize poorly. In this case, tuning only Mem or Dec layers yields suboptimal results (e.g., --17.8\% and --11.6\%), while tuning both layers gives a relative improvement of +4.7\%. This indicates that LCS provides site-aware signals which must be supported by joint adaptation of both low-level decoding (Dec) and higher-order memory mechanisms (Mem).
Overall, the best performance is achieved when both layers are fine-tuned together, benefiting from the synergy between site-specific representation (enabled by LCS) and task-level interaction (facilitated by LHA). This highlights the necessity of co-adapting spatially localized decoding and temporally enriched memory.

\begin{table}[t]
\centering
\resizebox{\columnwidth}{!}{
\begin{tabular}{lccccc|c}
\toprule
Indicator & Endo17. & Endo18. & Auto. & SCARED. & Stereo. & $\Delta m$ \\
\midrule
Random   & 3.08 & 1.85 & 0.82 & -7.68  & -3.20 & -1.03 \\
One-hot  & 2.55 & 1.98 & 0.88 & 12.02  & 0.25  & 3.53  \\
Ours & \textbf{5.23} & \textbf{2.12} & \textbf{1.13} & \textbf{18.42} & \textbf{2.73} & \textbf{5.92} \\
\bottomrule
\end{tabular}
}
\caption{Improvement (\%) of indicator designs in LCS.}
\label{lcs_indicator}
\end{table}

\begin{table}[t]
\centering
\resizebox{\columnwidth}{!}{
\begin{tabular}{lccccc|c}
\toprule
Indicator & Endo17. & Endo18. & Auto. & SCARED. & Stereo. & $\Delta m$ \\
\midrule
Random   & 2.48 & -3.99 & 0.95 & -8.08  & 2.31 & -1.26 \\
One-hot  & 2.50 & -3.33 & 1.03 & 6.36  & -0.15  & 1.28  \\
Ours & \textbf{5.23} & \textbf{2.12} & \textbf{1.13} & \textbf{18.42} & \textbf{2.73} & \textbf{5.92} \\
\bottomrule
\end{tabular}
}
\caption{Improvement (\%) of indicator designs in LHA.}
\label{lha_indicator}
\end{table}

\noindent \textbf{Efficiency and Communication Cost.} To quantify the overhead introduced by our method, Table~\ref{tab:combined_efficiency} reports the trained parameters, server-side parameters, and inference efficiency. Compared to baselines, our method introduces a negligible increase in trained parameters (22.35MB vs. 22.10MB) due to the LCS module, and a lightweight server-side module (0.25MB) for LHA. Importantly, only the trained parameters are transmitted during communication, ensuring efficiency comparable to existing methods. 
Regarding inference, while real-time performance is challenging for the SAM2 backbone on an NVIDIA RTX A5000, our LHA module does not affect latency as it is restricted to server-side aggregation. The LCS module introduces only minor overhead, with SurgFed achieving 0.36 FPS compared to 0.50 FPS for baselines, demonstrating that our enhanced multi-task adaptation comes at a minimal cost to deployment efficiency.

\section{CONCLUSIONS}

In this paper, we proposed a novel MTFL framework for surgical video understanding, termed as SurgFed, which tackles tissue and task diversity challenges faced by unique characteristics of surgical data. Specifically, we designed a novel LCS module by utilizing pre-defined text to incorporate surgical domain knowledge. Further, a LHA module was introduced to model the task interaction across different sites. Our method was extensively evaluated on five public datasets across four different surgical types and consistently outperformed state-of-arts methods.

\bibliographystyle{IEEEtran}
\bibliography{mybib}

\end{document}